\documentclass[conference]{IEEEtran}
\IEEEoverridecommandlockouts
\usepackage{cite}
\usepackage{amsmath,amssymb,amsfonts}
\usepackage{algorithmic}
\usepackage{graphicx}
\usepackage{textcomp}
\usepackage{xcolor}
\usepackage{float}
\usepackage{mathtools}
\usepackage{subfigure}
\usepackage{soul}
\usepackage{multirow,color,todonotes}
\usepackage{array} 
\usepackage{tikz,balance}
\usepackage{url}
\usepackage{enumitem,xcolor,booktabs,colortbl}
\usepackage[linesnumbered, ruled]{algorithm2e}
\SetKwRepeat{Do}{do}{while}
\def\BibTeX{{\rm B\kern-.05em{\sc i\kern-.025em b}\kern-.08em
    T\kern-.1667em\lower.7ex\hbox{E}\kern-.125emX}}

\begin{document}

\title{Generating Redstone Style Cities in Minecraft} 
\author{\IEEEauthorblockN{Shuo Huang\IEEEauthorrefmark{1}, Chengpeng Hu\IEEEauthorrefmark{1}, Julian Togelius\IEEEauthorrefmark{2}, Jialin Liu\IEEEauthorrefmark{1}}
 \IEEEauthorblockA{\IEEEauthorrefmark{1}
 \textit{Guangdong Provincial Key Laboratory of Brain-Inspired Intelligent Computation, }\\
 \textit{Department of Computer Science and Engineering, Southern University of Science and Technology, Shenzhen 518055, China}\\
 \IEEEauthorrefmark{2}
 \textit{Department of Computer Science and Engineering}\\
 \textit{Tandon School of Engineering, New York University, New York 10012, United States}}
 \thanks{This paper has been accepted by the 2023 IEEE Conference on Games.}
 }



\maketitle


\begin{abstract}
Procedurally generating cities in Minecraft provides players more diverse scenarios and could help understand and improve the design of cities in other digital worlds and the real world. This paper presents a city generator that was submitted as an entry to the 2023 Edition of Minecraft Settlement Generation Competition for Minecraft. The generation procedure is composed of six main steps, namely vegetation clearing, terrain reshaping, building layout generation, route planning, streetlight placement, and wall construction. Three algorithms, including a heuristic-based algorithm, an evolving layout algorithm, and a random one are applied to generate the building layout, thus determining where to place different redstone style buildings, and tested by generating cities on random maps in limited time. Experimental results show that the heuristic-based algorithm is capable of finding an acceptable building layout faster for flat maps, while the evolving layout algorithm performs better in evolving layout for rugged maps. A user study is conducted to compare our generator with outstanding entries of the competition's 2022 edition using the competition's evaluation criteria and shows that our generator performs well in the \textit{adaptation} and \textit{functionality} criteria.
\end{abstract}


\section{Introduction}
Minecraft~\cite{duncan2011minecraft} is a 3D sandbox video game widely used in research and education~\cite{nebel2016mining}. A well-built city in Minecraft can provide players a better game experience. Recent works focus on procedurally generating terrain and buildings in Minecraft \cite{awiszus2021world,jiang2022learning}. 
In the work of \cite{awiszus2021world}, generative adversarial network (GAN) is applied to generating terrain with similar features on Minecraft maps of different sizes. 
Jiang \textit{et al.}~\cite{jiang2022learning} employ procedural content generation via reinforcement learning (PCGRL)~\cite{khalifa2020pcgrl} to generate 3D mazes in Minecraft. 

The Settlement Generation Competition for Minecraft (GDMC)~\cite{salge2018generative} is a city generation competition. Participants are tasked to design a program that takes a Minecraft (version 1.19.2) map and a build area as the input and outputs a version of the same map with a city (or village) on it. The referees will assess each settlement with four criteria, namely \emph{adaptation}, \emph{functionality}, \emph{believable and evocative narrative}, and \emph{visual aesthetic} as described in \cite{salge2018generative}. The challenges of settlement building in Minecraft are discussed in \cite{salge2020ai,salge2019generative}. Green \emph{et al.} \cite{green2019organic} generate buildings with constrained growth and cellular automata, while Barthet \emph{et al.} \cite{barthet2022open} use an evolutionary algorithm. Merino \emph{et al.} \cite{merino2023interactive} generate buildings based on individual aesthetic choice and interactive evolution. In \cite{clappers2021simulation}, a needs-based agent is presented that simulates the development process of a village. Demke \emph{et al.} \cite{demke2021multi} design a multi-agent story-based settlement generator.

Following the rules of GDMC, this paper designs a city generator using redstone buildings (i.e., buildings with inner redstone designs) as components. Redstone is a mineral that can transit redstone signals, which can be used to simulate electrical signals, between blocks. Redstone design makes the city show a scientific style.
Three building layout generation algorithms are integrated into our generator and compared by generating cities on maps with different sizes in limited time. A user study is conducted to compare our work with some outstanding ones from the 2022 edition of GDMC. 

\section{Generation Process}
\begin{figure*}[htb]
    \centering
    \includegraphics[width=.9\textwidth]{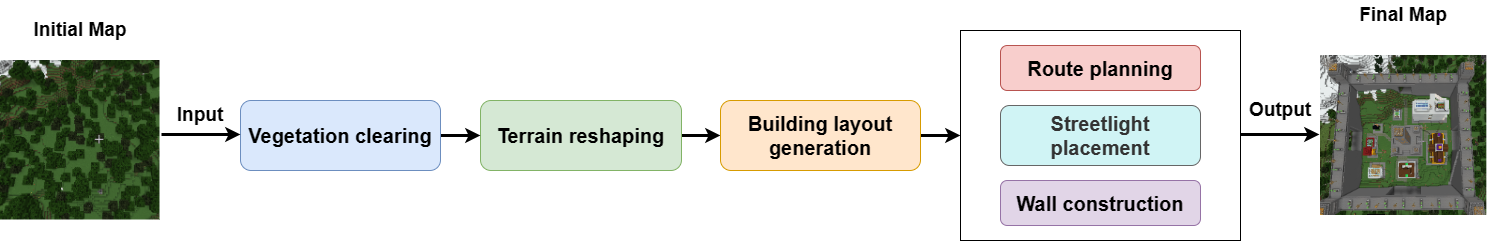}
    \caption{Main steps of our city generator.}
    \label{ProcessPicture}
\end{figure*}

The city generation process consists of six main steps: vegetation clearing, terrain reshaping, building layout generation, route planning, streetlight placement, and wall construction as illustrated in Fig. \ref{ProcessPicture}.
First, we clear the trees and reshape the terrain to get more building area. Then, the whole city is divided into an outer and an inner part. The outer part is used for constructing walls and the inner part is left for placing buildings. 
After determining the locations of buildings, some important nodes are connected by roads to provide more accessible connections among different parts of the city. Finally, some streetlights are placed in the inner city to offer enough lighting. 
The following details the steps.

{\textbf{Vegetation clearing:}}
In some maps, trees occupy plenty of space. As enough building space is necessary for placing buildings, our generator clears all the trees.

{\textbf{Terrain reshaping:}}
The ideal area for placing buildings is a flat area, where all the blocks have about the same elevation. Therefore, we use a rule-based strategy to reshape the terrain (cf. Algorithm \ref{reshapeAlgorithm}).

\begin{algorithm}[htbp]
\caption{Terrain reshaping }\label{reshapeAlgorithm}
\KwData{Width of the building area $w$, Length of the building area $l$, Most common block of the building area's ground $gb$, Function used to compare two blocks' altitude $compareAltitude$}

 $A$ $\leftarrow$ building area's altitude map\\
mean $\leftarrow$ average of $A$\\
\For{x = 1 to w}{
    \For{z = 1 to l}{
        score $\leftarrow$ $compareAltitude$($x$,z)\\
        \If{score = 3}{
            change block in ($x$,$A$[$x$][$z$]-1,$z$) to air\\
            change block in ($x$,$A$[$x$][$z$]-2,$z$) to $gb$
        } 
        \If{score = -3}{
            change block in ($x$,$A$[$x$][$z$],$z$) to $gb$
        } 
    }
}
\Return{buildArea}
\end{algorithm}


{\textbf{Building layout generation:}
After preserving enough empty place, the generator determines the layout of a set of redstone style buildings pre-designed. 
Some buildings involve some redstone design as shown in Fig. \ref{redStoneStructure}.

The layout generation is modelled as an optimization problem.
The terrains are categorized into 5 types, including plain, common land, water, artificial blocks, and blocks around the artificial blocks. The costs and rewards of placing buildings on different terrain type are shown in Table \ref{cost} and Table \ref{areaTable}.  
The Monument offers an additional bonus: the closer it is to the city's center point, the greater the reward is. There is a minimum distance between each pair of buildings. The reward after deducting the cost is the score for placing a building.

\begin{table}[htbp]
    \begin{center}
        \caption{Building cost on different type of terrain.}\label{cost}
        \begin{tabular}{ll} 
            \toprule
            Category & Cost \\
            \midrule
            Water & 10\\
            Plain & 0\\
            Artificial blocks & 10000 \\
            Blocks around the artificial & 50 \\
            Common land & The sum of its gradient \\
            \bottomrule
        \end{tabular}
    \end{center}
\end{table}

\begin{table}[htbp]
    \begin{center}
        \caption{Building rewards for  different buildings}\label{areaTable}
        \begin{tabular}{ll} 
            \toprule
            Building & Reward \\
            \midrule
            Dorm & 493\\
            Church & 416\\
            Munition Factory & 119\\
            Monument & 625\\
            Shop & 182\\
            HIM Statue & 35\\
            Enderman Statue & 25\\
            Trampoline & 35\\
            Enchanting Room & 121\\
            \bottomrule
        \end{tabular}
    \end{center}
\end{table}

 
  

\begin{figure}[htbp] 
	\centering  
	\subfigtopskip=2pt 
	\subfigbottomskip=2pt 
	\subfigcapskip=-5pt 
 
	\subfigure[Enchanting table]{
		\label{ect}
		\includegraphics[width=0.29\columnwidth]{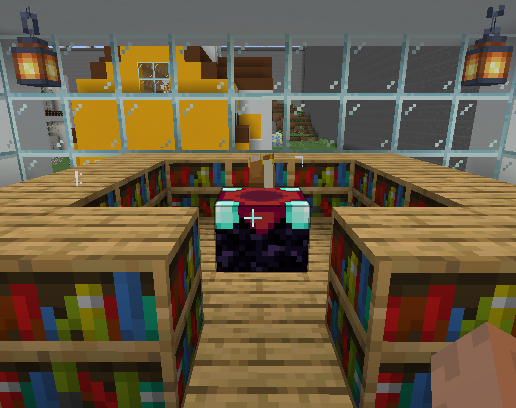}}
	\quad 
	\subfigure[Light]{
		\label{light}
		\includegraphics[width=0.34\columnwidth]{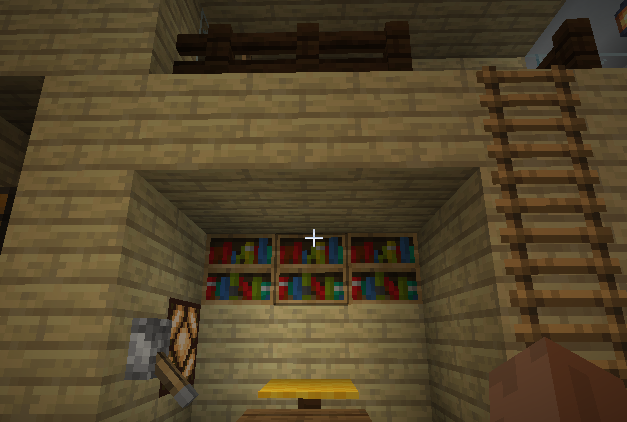}}
	\quad
	\subfigure[Farm]{
		\label{farm}
		\includegraphics[width=0.21\columnwidth]{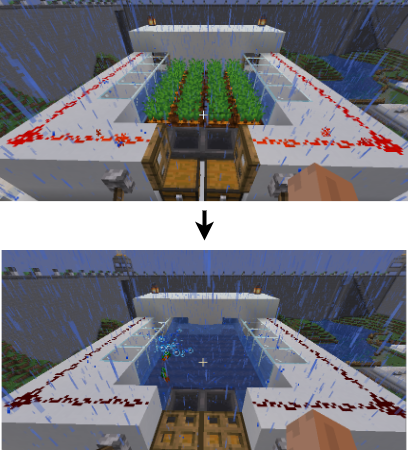}}
	\quad
	\caption{Redstone designs.}
	\label{redStoneStructure}
\end{figure}

Search-based algorithms can be applied to optimize the layout aiming at maximizing the reward. Three algorithms are tested, namely a heuristic-based layout generation algorithm, an evolving layout algorithm, and random placement.
Algorithm \ref{stepRandomAlgorithm} details the heuristic-based layout generation process. It generates many layouts heuristically and returns the one with the highest score. When generating a layout, it adds buildings one by one until the score is no more increased by adding new buildings. 
Algorithm \ref{evoAlgorithm} details the process of evolving layout. In an $m*n$ map, every position  $(x, z)$ can be represented by a positive integer $k=x*n+z$. The number of buildings needed to be placed, $num$, can be input by users. A solution $s$ is made up of $num$ positive integers. If a solution breaks the rule of minimum distance between two buildings, only one of them would be placed on the map. A solution can be transformed into a layout. The fitness is the total score of the layout. The initial population is generated randomly. Random placement algorithm uses $num$ random numbers to represent the positions of buildings in one layout. Many layouts would be generated and the best one would be chosen.

\begin{algorithm}[htbp]
\caption{Heuristic-based layout generation.}\label{stepRandomAlgorithm}
\KwData{Number of layouts need to be generated $nb$, Maximum times can be used to add one building to a layout $maxTry$}

$A$ $\leftarrow$ the altitude map of the inner city\\

\For{i = 1 to nb}{
    tempReward $\leftarrow$ 0\\
    layout $\leftarrow$ $A$\\
    maxScore $\leftarrow$ 1\\
    \While{maxScore$>$0}{
        maxScore $\leftarrow$ -1\\
        \For{j = 1 to $maxTry$}{
            $c$ $\leftarrow$ copy of layout\\
            randomly place a building $b$ on $c$\\
            score $\leftarrow$ getReward ($b$) - getCost ($b$)\\
            update maxScore\\
        }
        update layout\\
        tempReward $\leftarrow$ tempReward + maxReward\\
    }
    update bestReward\\
    update globalLayout\\
}
\Return{globalLayout}
\end{algorithm}

\begin{algorithm}[htbp]
\caption{Evolving layout.}\label{evoAlgorithm}
\KwData{Size of population $popSize$, Evolutionary generation $NGeneration$, Mutation rate $pumt$}
$A$ $\leftarrow$ the altitude map of the inner city\\
pop $\leftarrow$ initial population\\
fitnessList $\leftarrow$ []\\
newPop $\leftarrow$ []\\
\For{ng = 1 to NGeneration}{
    \For{i = 1 to popSize}{
        layoutCopy $\leftarrow$ $A$\\
        fitnessList[i] $\leftarrow$ calculateFitness(pop[i], layoutCopy)
    }
    parents $\leftarrow$ selectParents(pop, fitnessList, popSize)\\
    \For{i = 1 to popSize}{
        Randomly selection parents x, y\\
        child $\leftarrow$ reproduce(x, y)\\
        $p$ $\leftarrow$ a random number between 0 and 1.\\
        \If{p $<$ $pumt$}{
            child $\leftarrow$ mutate(child)
        }
        update(newPop)
    }
    pop $\leftarrow$ newPop
}
bestSol $\leftarrow$ the best solution in pop\\
globalLayout $\leftarrow$ bestSol's corresponding layout

\Return{globalLayout}
\end{algorithm}


{\textbf{Route planning:}}
After determining the locations of buildings, roads are planned to ensure the accessibility of important areas. Four gates are placed at the closest legal points to the middle positions of four city edges. Roads should connect the four gates and the monument. In addition, the munitions factory is connected to the tower's entrance. 
All the roads are planned by an A$^*$ algorithm, in which the sum of two points' difference in $x$ and $z$ values is used to estimate their distance. Finally, roads on the water are transformed into bridges.

{\textbf{Streetlight placement:}}
Streetlights need to be placed in the area with high building density to provide lighting to the city.  KMEANS~\cite{macqueen1967some} is used to find the most suitable locations for streetlights. However, the returned locations are not always legal. Therefore, breadth first search is applied to the output of KMEANS to determine the final locations of streetlights.

{\textbf{Wall construction:}}
A wall, consisting of the wall base, plane and tower, is built in the outer part of the city to protect the residents. The height of the wall base is adjusted to fit the terrain. Torches and cannons are evenly placed on the plane.

An example of our generator's output is shown in Fig. \ref{cityRes}.

\section{Experiments}
Two sets of experiments are designed to test the performance of our generator. One aims at comparing the three building layout generation algorithms in different maps, and the other performs a user study, in which our city generator is compared with two outstanding ones from the 2022 edition of the GDMC. 

{\textbf{Comparison of layout generation algorithms:}}
The algorithms are tested on the four maps described in Table \ref{maps}. 
The total score of placing buildings according to the generated layout is computed to assess the corresponding algorithm. 
The population size, mutation rate and generation number in Algorithm \ref{evoAlgorithm} are set as $PopSize=100$, $Pmut=0.1$ and $NGeneration=100$, respectively.
Each algorithm is run 30 times. Fig. \ref{compare1} shows the convergence curves.

\def\nouse{
\begin{table}[htb]
    \begin{center}
        \caption{Parameter setting}\label{parameters}
        \begin{tabular}{lll} 
            \toprule
            Name & Description & Value \\
            \midrule
            PopSize & Size of population & 100\\
            NGeneration & Evolutionary generation  & 100\\
            Pmut & Mutation rate & 0.1 \\
            \bottomrule
        \end{tabular}
    \end{center}
\end{table}}

\begin{table}[htb]
    \begin{center}
        \caption{Map message}\label{maps}
        \begin{tabular}{cccc} 
            \toprule
            Map & Map size & \#Buildings & Ratio of plain \\
            \midrule
            (a) & 100 & 5 & 60.6\%\\
            (b)  & 100 & 5 & 48.2\%\\
            (c) & 150 & 9 & 53.9\%\\
            (d) & 150 & 9 & 25.2\%\\
            \bottomrule
        \end{tabular}
    \end{center}
\end{table}

\begin{figure}[ht] 
	\centering  
	\subfigtopskip=2pt 
	\subfigbottomskip=2pt 
	\subfigcapskip=-5pt 
	\subfigure[Flat 100*100 map]{
		\label{flat}
		\includegraphics[width=0.4\linewidth]{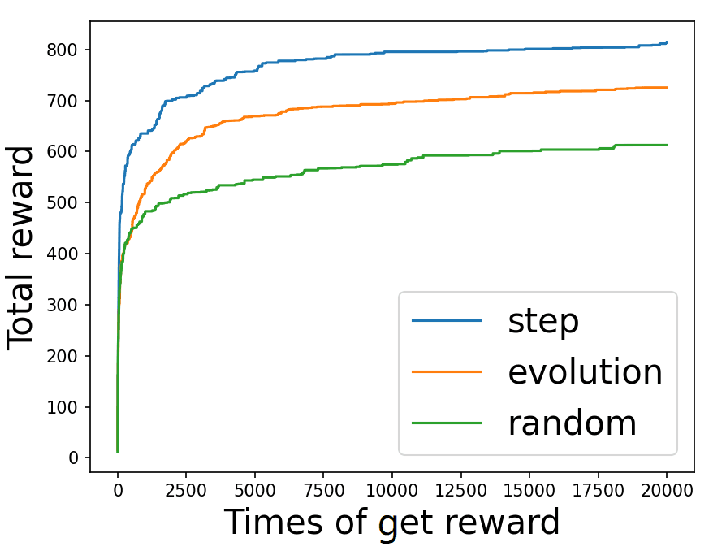}}
	\quad 
	\subfigure[Rugged 100*100 map]{
		\label{cliffy}
		\includegraphics[width=0.4\linewidth]{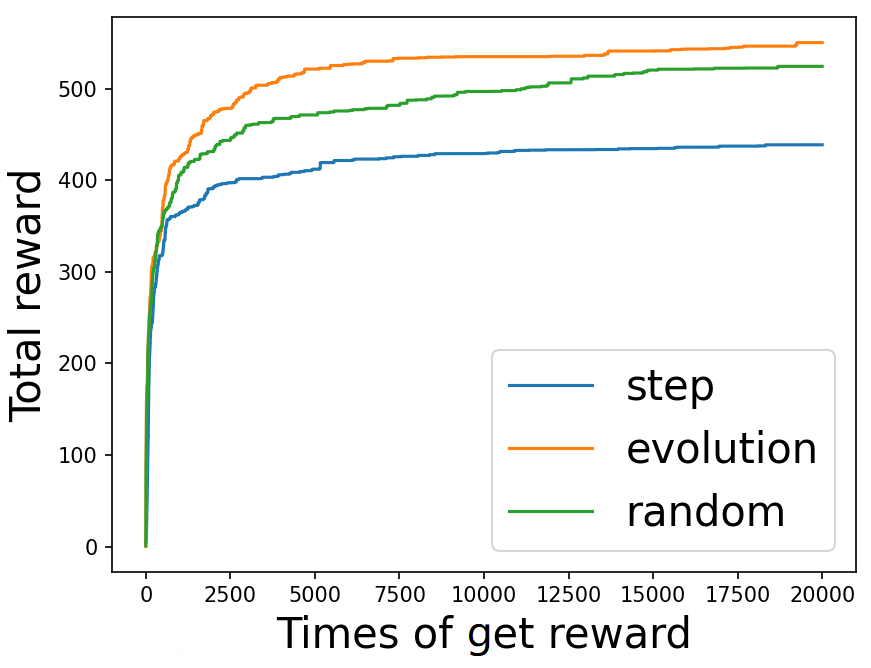}}
\\
	\subfigure[Flat 150*150 map]{
		\label{flat_2}
		\includegraphics[width=0.4\linewidth]{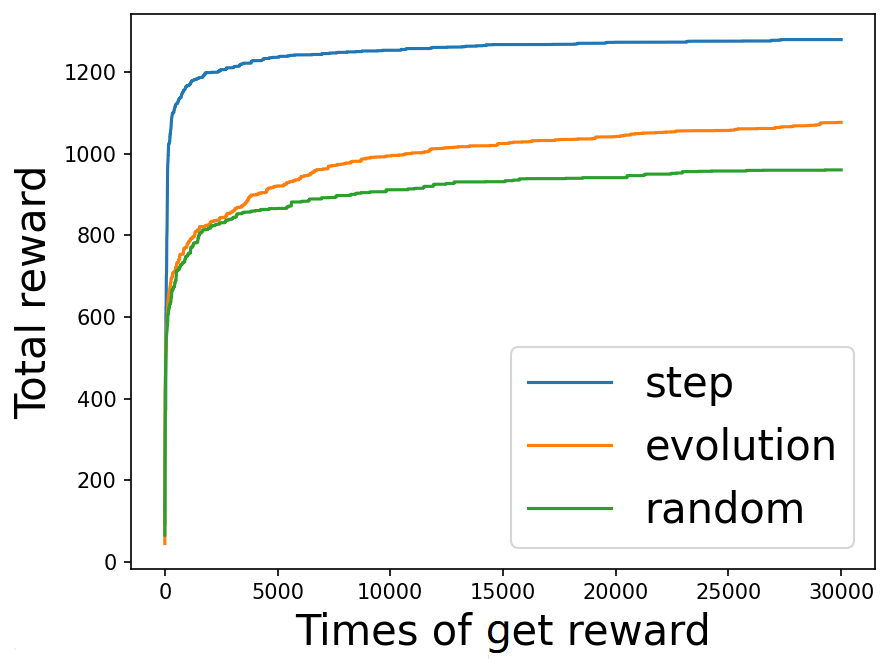}}
	\quad
	\subfigure[Rugged 150*150 map]{
		\label{cliffy_2}
		\includegraphics[width=0.4\linewidth]{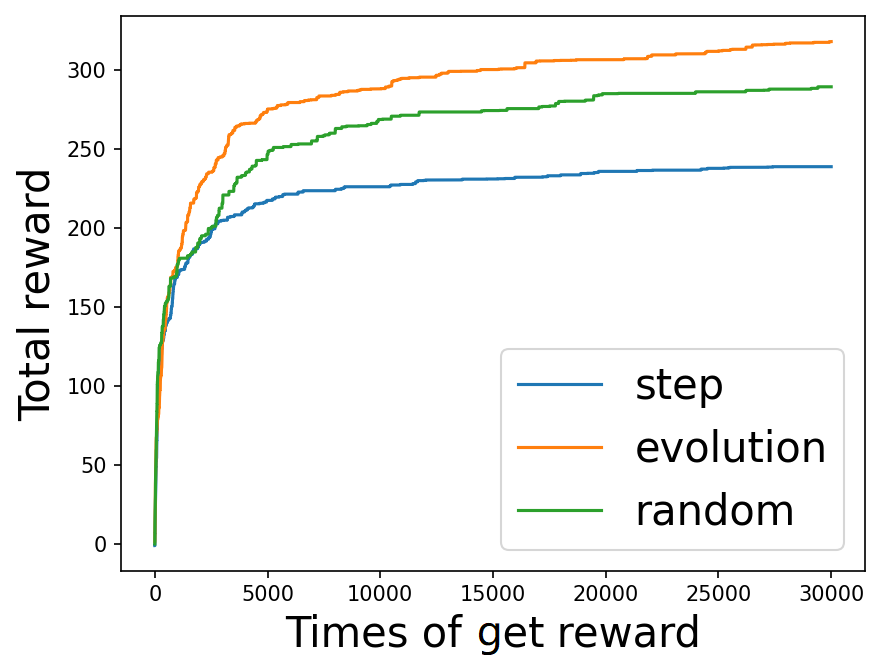}}
	\caption{Comparison among three algorithms on maps with different sizes.}
	\label{compare1}
\end{figure}

In flat maps like (a) and (c), the heuristic-based layout generation algorithm can find a layout with an acceptable score quickly. It may be explained by the fact that this algorithm guarantees the legality of layout, thus there is no collision among buildings. However, it does not use previously generated layouts well, and performs poorly in some rugged maps like (b) and (d). In contrast, the evolving layout algorithm can generate better new layouts based on the good ones in the population, which significantly increases the efficiency.

\begin{figure}[htbp]
    \centering
    \includegraphics[width=.2\textwidth]{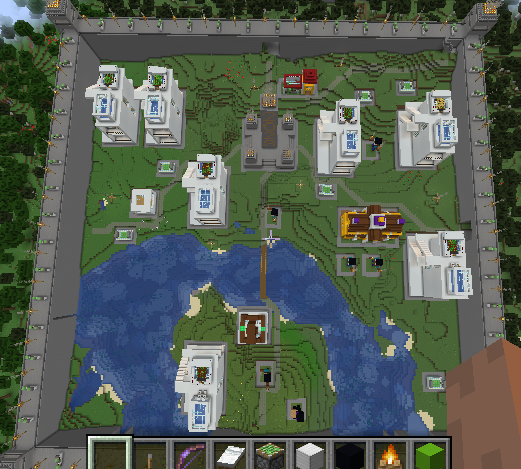}
    \hspace{1em}
    \includegraphics[width=.19\textwidth]{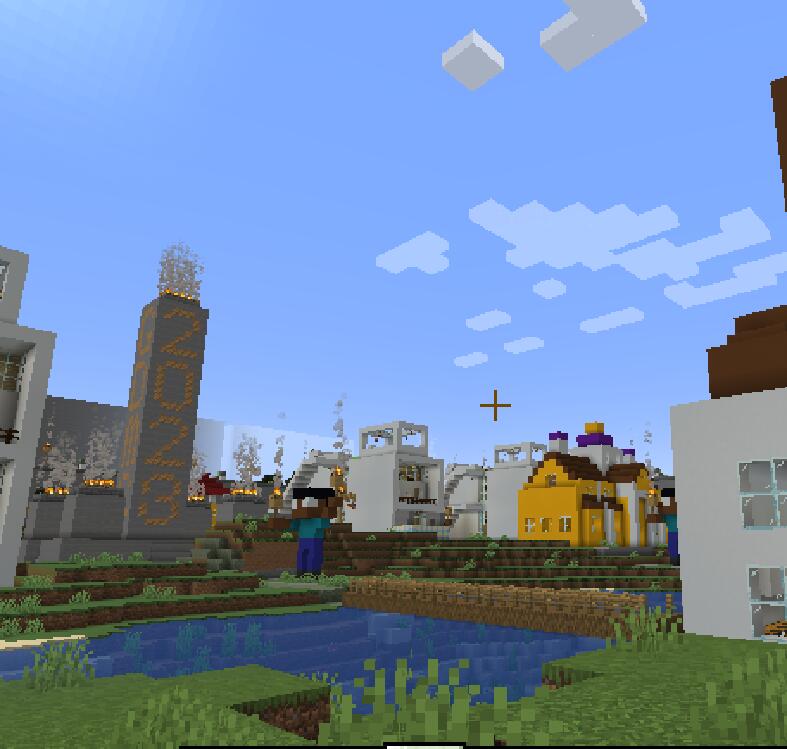}
    \caption{Final output}\label{cityRes}
\end{figure}

{\textbf{{User study:}}
A user study is conducted following the evaluation criteria of GDMC to evaluate the performance of our generator. Six undergraduate and graduate students have participated. All of them are between 21 – 26 years old. They all have some basic knowledge of Minecraft.
Our generated city is compared with two entries of GDMC'2022: \textit{PandaVision} and \textit{Medieval City} found on the competition website (\url{https://gendesignmc.engineering.nyu.edu/results}). \textit{Medieval City} is the one with the highest overall score in GDMC'2022. \textit{PandaVision} offers a detailed documentation which facilitate its usage.

Every participant needs to visit each city generated by the three works for at least five minutes in a first view.
They are asked to rank the three generators with a score of 1-10 based on their performance on four criteria, \emph{adaptation}, \emph{functionality}, \emph{believable and evocative narrative}, and \emph{visual aesthetic} as in the competition's official evaluation process.
Table \ref{userStudy} reports the user study with six participants.

Result shows that our generator performs better in adaptation, probably because it only reshapes the terrain slightly and considers the influence of terrain a lot. \textit{PandaVision} generator flattens the entire building area and lays the stone slab directly, which looks out of place in the environment. 
The \textit{Medieval City} generator performs badly on functionality. It does not pay enough attention to the accessibility of the city. Some ladders cannot be climbed, and some paths cannot be passed.
Our generator outputs maps with poor narrative. That may be blamed to the hidden redstone structures in the buildings, that are hard to be seen by users.

\begin{table}[htb]
    \begin{center}
        \caption{Result of user study.}\label{userStudy}
        \begin{tabular}{lccc} 
            \toprule
            Criteria & Redstone Technique & PandaVision & Medieval City \\
            \midrule
            Adaptation & 7.67 & 6.33 & 7.50 \\
            Functionality & 7.50 & 7.33 & 6.67\\
            Narrative & 7.83 & 8.67 & 7.67\\
            Aesthetic & 7.33 & 8.50 & 7.50\\
            \midrule
            Overall & 7.58 & 7.71 & 7.33\\
            \bottomrule
        \end{tabular}
    \end{center}
\end{table}

\section{Conclusion}
This paper designs a redstone style city generator to create a city on given Minecraft maps, as a competition entry to the GDMC'2023. Three algorithms are used to determine the position of buildings and are compared on maps with different sizes and terrain types. Our generator is compared with two representative entries of the GDMC'2022 and shows competitive performance according to user assessment. 
As future work, it is worth integrating building generation techniques (such as \cite{green2019organic,barthet2022open,merino2023interactive}) to increase the generator's performance on \textit{Narrative} and \textit{Aesthetic} criteria while minimizing the human input for designing buildings. More professional participants will be invited to evaluate the generator comprehensively.

\bibliographystyle{IEEEtran} 
\bibliography{minecraft}
\end{document}